\documentclass[conference]{IEEEtran}
\def\BibTeX{{\rm B\kern-.05em{\sc i\kern-.025em b}\kern-.08em
    T\kern-.1667em\lower.7ex\hbox{E}\kern-.125emX}}

\usepackage{graphics} 
\usepackage{epsfig} 
\usepackage{mathptmx} 
\usepackage{times} 
\usepackage{amsmath} 
\usepackage{amssymb}  
\usepackage{color}
\usepackage{soul}
\usepackage{siunitx}
\usepackage{multirow}
\usepackage{graphicx}
\usepackage{float}
\usepackage{cite}
\usepackage{soul}
\usepackage{booktabs}
\usepackage{multirow}
\usepackage{hyperref}
\usepackage{caption}
\newcommand{\update}[1]{{\color{black}{#1}}}

\title{\LARGE\bf \update{Curvature-Aware Calibration of Tactile Sensors for Accurate Force Estimation on Non-Planar Surfaces}}

\author{
\IEEEauthorblockN{
\begin{minipage}{0.50\textwidth}
\centering
Luoyan Zhong\\
\textit{School of Electrical and Computer Engineering} \\
\textit{Cornell University}\\
Ithaca, NY, USA\\
lz572@cornell.edu
\end{minipage}
\hfill
\begin{minipage}{0.50\textwidth}
\centering
Heather Jin Hee Kim\\
\textit{Sibley School of Mechanical and Aerospace Engineering} \\
\textit{Cornell University}\\
Ithaca, NY, USA\\
jk2768@cornell.edu
\end{minipage}
}
\vspace{1em}
\IEEEauthorblockN{
\begin{minipage}{0.50\textwidth}
\centering
Dylan P. Losey\\
\textit{Dept. of Mechanical Engineering} \\
\textit{Virginia Tech}\\
Blacksburg, VA, USA\\
losey@vt.edu
\end{minipage}
\hfill
\begin{minipage}{0.50\textwidth}
\centering
Cara M. Nunez\\
\textit{Sibley School of Mechanical and Aerospace Engineering} \\
\textit{Cornell University}\\
Ithaca, NY, USA\\
cmn97@cornell.edu
\end{minipage}
}
}

\begin{document}

\maketitle

\begin{abstract}
Flexible tactile sensors are increasingly used in real-world applications such as robotic grippers, prosthetic hands, wearable gloves, and assistive devices, where they need to conform to curved and irregular surfaces.
However, most existing tactile sensors are calibrated only on flat substrates, and their accuracy and consistency degrade once mounted on curved geometries. This limitation restricts their reliability in practical use.
To address this challenge, we develop a calibration model for a widely used resistive tactile sensor design that enables accurate force estimation on one-dimensional curved surfaces. We then train a neural network (a multilayer perceptron) to predict local curvature from baseline sensor outputs recorded under no applied load, achieving an $R^{2}$ score of 0.91.
The proposed approach is validated on five daily objects with varying curvatures 
under forces from 2 N to 8 N. 
Results show that the curvature-aware calibration maintains consistent force accuracy across all surfaces, while flat-surface calibration underestimates force as curvature increases. 
Our results demonstrate that curvature-aware modeling improves the accuracy, consistency, and reliability of flexible tactile sensors, enabling dependable performance across real-world applications.

\end{abstract}

\begin{IEEEkeywords}
tactile sensing, flexible sensor, curvature, calibration model
\end{IEEEkeywords}

\section{Introduction}

Tactile sensing is an essential component in robotics and wearable systems, enabling machines and humans to perceive and interact safely with their environment. Current applications range from prosthetic hands and robot grippers to wearable gloves and assistive technologies, where tactile feedback is used to estimate parameters such as contact force, pressure, or texture.

To conform to complex and often curved surfaces, tactile sensors are often designed using flexible or stretchable materials. Typical examples include elastomer-based composites, polymer films, and textile substrates integrated with conductive elements such as carbon or silver-based inks, liquid metals, or conductive threads~\cite{dahiya_tactile_2010, wan_recent_2017, pyo_recent_2021}. These compliant designs allow sensors to be attached to diverse geometries—from flat plates to curved robot bodies or human skin—while maintaining good mechanical conformity.

However, the same material properties that provide flexibility also make these sensors sensitive to the geometry of the surface to which they are attached. When bent or mounted on irregular surfaces, the local strain distribution and layer compression change, leading to geometry-dependent signal variations that can distort the relationship between the measured electrical response and the true applied force~\cite{wang2017flexible, khodasevych2017flexible}. Despite this well-known issue, most prior work has not systematically addressed it. Some researchers have calibrated their sensors only on flat surfaces and then directly extended the results to curved applications~\cite{xu_cushsense_2024, fiedler2021low}; others have calibrated on a specific surface curvature, making the resulting models valid only for that single geometry~\cite{verma2025flexible}. A few have noted the curvature effect qualitatively but lacked a quantitative correction method~\cite{si_robotsweater_2023}. As a result, the sensing accuracy of flexible tactile sensors often degrades in real-world settings where surface curvature varies.

To address this challenge, we introduce a lightweight flexible tactile sensor system together with a calibration model that explicitly accounts for the curvature of the surface to which the sensor is attached. Our sensor predicts one-dimensional curvature on rigid objects using a trained neural network and provides calibrated force readings. By integrating curvature-aware calibration, our approach enables accurate estimation of static forces across non-planar surfaces, thereby improving sensing fidelity in applications where surface geometry cannot be assumed to be flat. To ensure accessibility, both the hardware and software are open-source and can be easily reproduced
: \url{github-link-anonymized-for-initial-submission}. 


\begin{figure}[t]
    \centering
    \includegraphics[width=1.0\linewidth]{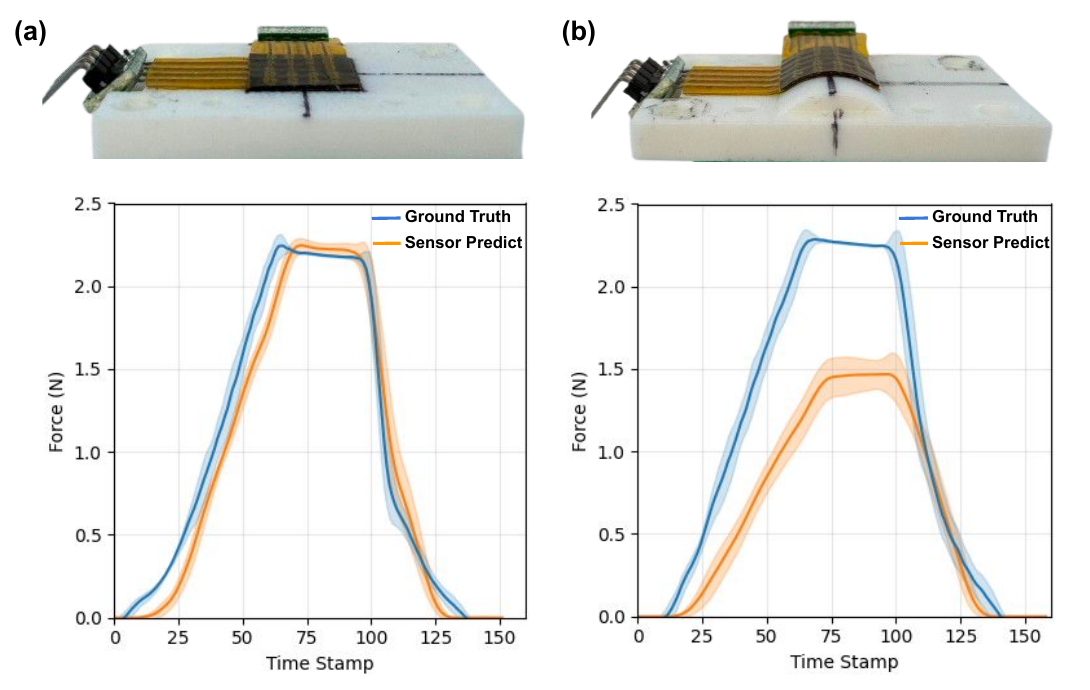}
    \caption{\textbf{Predicted Force Response on Flat and Curved Surfaces.} (a) On a flat surface, the calibrated sensor predicts forces that closely match the ground truth. (b) On a curved surface, using the same flat calibration leads to significant deviation from the ground truth, highlighting the need for curvature-aware calibration.}
    \label{fig:sensor}
\end{figure}
\section{Related Work}

\subsection{Tactile Sensor Designs} 



Tactile sensors are crucial wherever manipulators physically interact with their environment, playing especially vital roles in human–machine interfaces, robotics, teleoperation, prosthetics, and wearable devices. In these applications, the tactile sensors require flexible and conformable properties in order to cover large and irregular surfaces~\cite{wang_recent_2023, jin_progress_2023}. A variety of sensing modalities have been explored to achieve these properties, including resistive, capacitive, piezoelectric, and magnetic arrays. Capacitive and piezoelectric designs can deliver high sensitivity and dynamic response, but typically require multilayer stacks, shielding, and specialized readout circuits~\cite{xu_flexible_2025,luo_sock-embedded_2023,yu_simple_2023,huang_flexible_2021},\cite{ku_cmos-integrated_2025,zhen_high-density_2023,nair_transparent_2023,seminara_piezoelectric_2013,khan_flexible_2015}. Magnetic tactile sensors, which combine small magnets with Hall-effect elements, are increasingly adopted for multi-axis force estimation, yet scaling them to large-area arrays remains challenging~\cite{li_high-sensitivity_2024,liu_novel_2025,ji_magnetic_2025,zhang_design_2024}.

In contrast, resistive arrays offer low cost, simple fabrication, and scalability, which makes them suitable for a wide range of applications, including large-area robotic skins, prosthetics, switch buttons, and touch displays \cite{stassi2014flexible, zhu2022recent}. These arrays are typically implemented by patterning electrodes into a row–column matrix and embedding a pressure-sensitive conductive film as the sensing layer. Commonly used materials include carbon-loaded polymer sheets, such as Velostat~\cite{liu2017glove, suprapto2017low}, and fabric-based composites, such as EeonTex~\cite{pannen2022low}. These materials are inexpensive, flexible, and conformable to curved substrates. In practice, electrodes are often realized with copper tape or conductive thread, enabling straightforward prototyping and scalability. However, the response of these resistive films is also influenced by deformation such as bending or pre-strain, which can introduce variability into the signals \cite{fatema_investigation_2022, tihak_experimental_2019}. This limitation highlights the need for improved calibration strategies and compensation methods that can enhance the accuracy of resistive tactile sensors when deployed on curved surfaces. 
To address this, we aim to employ a commonly used resistive sensor design, perform calibration, and evaluate the calibration against a range of everyday objects with non-planar shapes.

\subsection{Impact of Sensor Deformation Due to Object Curvature}

As described in the previous subsection, tactile sensors commonly leverage materials that undergo property changes under deformation, which serve as the physical basis for signal transduction in various sensing mechanisms. For capacitive sensors, deformation of the dielectric layer between conductive plates modulates the device capacitance. 
Resistive tactile sensors operate on piezoresistive principles, in which mechanical deformation alters microstructures or conductive pathways within the sensitive layer, resulting in measurable resistance changes. These changes can be used to detect mechanical states, such as strain or bending~\cite{somappa_3d_2020, harija_novel_2024}. However, decoupling sensor deformation due to applied contact compared to environmental factors, such as object curvature, remains a challenge.

More specifically, materials used for resistive tactile sensors are frequently polymer matrices embedded with conductive fillers such as carbon black, carbon fibers, carbon nanotubes, or graphene to achieve piezoresistive performance. When these compliant, resistive tactile sensors are applied to curved surfaces, bending introduces non-uniform strain profiles across the sensor thickness; compression occurs on one side while tension is present on the other, which directly impacts charge transport and thus sensor response. The effects of non-uniform strain become more pronounced in sensors with multiple conductive layers, potentially leading to compounded gradients and further impacting electrical properties and overall sensing performance. Figure~\ref{fig:sensor} illustrates this effect, showing that a sensor calibrated on a flat surface performs accurately on planar contact, but deviates significantly when mounted on a curved surface. 

In this work, we calibrate a resistive tactile sensor made with Velostat under varying curvatures and demonstrate an approach that can identify and decouple the impact of sensor deformation due to curvature.




\subsection{Calibration Models for Tactile and Force Sensing}


Calibration methods for resistive tactile sensors generally focus on linearization of force-resistance, often through machine learning models, finite element analysis, or empirical correction using reference instruments~\cite{ye2020inconsistency, hu2023machine, lee2020calibrating}. While these approaches improve accuracy on flat surfaces, they rarely incorporate curvature as an explicit parameter in the calibration process.

Exceptions exist in specialized domains, such as robotic manipulation or haptic rendering, where geometry-aware corrections have been explored~\cite{verma2025flexible}, but a generalizable curvature-aware calibration framework for flexible tactile sensors remains underdeveloped. This gap motivates the need for a dedicated calibration strategy that can compensate for the effects of surface curvature on force measurement fidelity. To address this, we employ easily 3D-printable hardware test rigs, a commonly available linear motor, and a force sensor (which can be substituted with other models) to build a force calibration pipeline that accounts for object curvature. 
\section{Hardware} \label{sec:hardware}

\begin{figure}[t]
    \centering
    \includegraphics[width=1.0\linewidth]{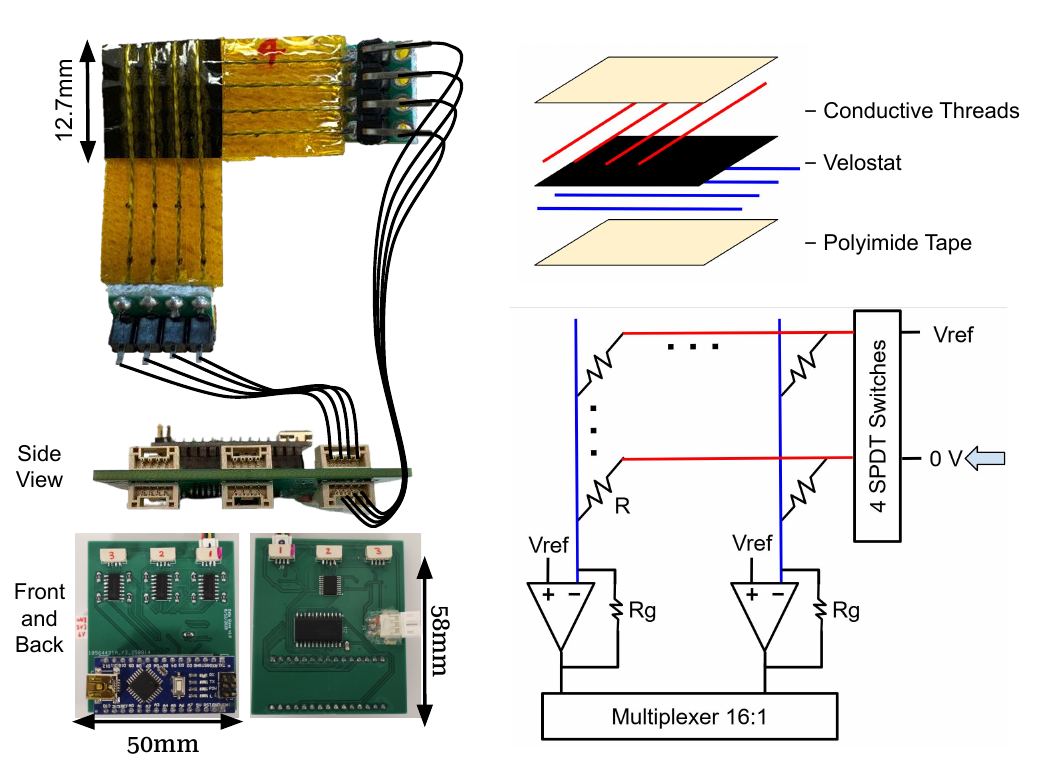}
    \caption{\textbf{Tactile Sensing Hardware.} The sensor was constructed by sandwiching a Velostat film between two orthogonal layers of conductive threads, forming a resistive array. A custom PCB handled row-column scanning, sequentially addressing each node and sending the measurements to an Arduino. The circuit schematic illustrates the scanning architecture.}
    \label{fig:hardware}
\end{figure}

In this section, we describe the design and implementation of our flexible tactile sensor. We present the construction of an individual sensing unit and the readout circuit used to acquire sensor signals, and introduce the experimental setup used to evaluate sensor behavior on surfaces with different curvatures. The materials used to construct our sensor are widely available, as is the fabrication process. 


\subsection{Sensor Fabrication}

The tactile sensor consists of five functional layers arranged in a stacked configuration (Fig. \ref{fig:hardware}). The middle layer is made of a piezoresistive sheet (Velostat, Adafruit), which forms the active sensing medium. Above and below the Velostat are layers of 2-ply conductive thread (Adafruit), taped in a grid pattern to serve as electrodes. The outermost layers consist of polyimide tape, which provides insulation and protects the sensor from shorting or mechanical wear. Due to the piezoresistive nature of Velostat and the cross-grid electrode layout, the complete sensor behaves as a $4\times4$ variable resistor array. When pressure is applied, the resistance at the corresponding node decreases in proportion to the applied force.  The fabricated sensor unit measures $0.5$~in~$\times$~$0.5$~in and contains 16 sensing nodes arranged in a $4\times4$ matrix. Each intersection of the top and bottom conductive thread electrodes defines a sensing node. 

 By combining the sensor with the readout circuit described in Section~\ref{sec: readout_circuit}, we obtain voltage outputs corresponding to the resistance changes at each node, which are then mapped to force values applied at specific locations. 

\subsection{Circuitry}
\label{sec: readout_circuit}

The $4\times4$ sensor array is addressed with a row-column scanning scheme. Each row is connected to a bank of single pole double throw (SPDT) analog switches, which sequentially drive one row to ground while holding all other rows at $V_{\text{ref}}$. This row-biasing ensures that only the sensing nodes on the active row contribute to the measurement, minimizing cross-talk between nodes.

Each column output is routed directly to a non-inverting amplifier, which isolates the sensor array and provides gain adjustment. The amplifier output is then passed through a $16{:}1$ analog multiplexer to a single analog to digical converter channel on the microcontroller. By stepping through the row select lines and column multiplexer addresses, the controller raster-scans all 16 nodes to reconstruct the pressure distribution.

The output voltage follows the standard non-inverting amplifier relation based on the circuit schematic in Fig.~\ref{fig:hardware}:
\vspace{-.3em}
\[
V_{\text{out}} = \Bigl(1 + \frac{R_g}{R}\Bigr)V_{\text{ref}}
\]
where $R_g = 5.6~\text{k}\Omega$ is the feedback resistor and $R$ is the pressure-dependent resistance of the Velostat layer. Since $R$ decreases under applied pressure, $V_{\text{out}}$ varies monotonically with force, enabling calibration to map readings to applied loads.

\section{Curvature Prediction}
\label{sec:curv}
We collected baseline readings (no force is applied) from all 16 sensing nodes on multiple curved surfaces; these baseline responses reflect how the sensor deforms when attached to a curved surface. We then formulate this as a supervised regression problem and design a residual multilayer perceptron (MLP) to map the sensor input array to a one-dimensional curvature value. For general 3D surfaces, we report the mean curvature $\kappa=\tfrac{1}{2}(k_1+k_2)$ at the contact patch. For the cylindrical fixtures used in training, this reduces to $\kappa=1/R$ since $k_2=0$.

\subsection{Architecture Description}
\begin{figure}[b]
    \centering
    \includegraphics[width=.8\linewidth]{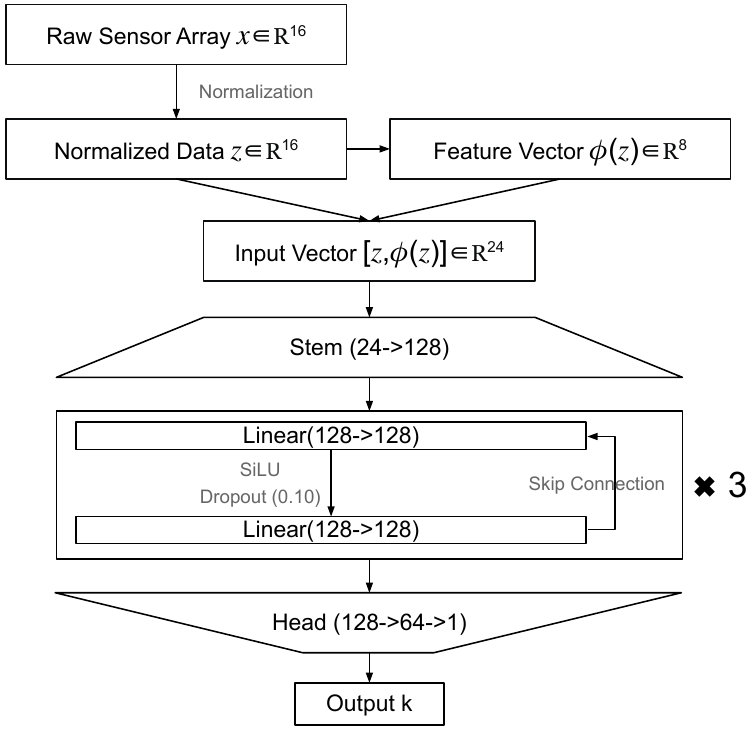}
    \caption{\textbf{MLP Network Architecture.} Residual multilayer perceptron that predicts surface curvature directly from baseline tactile sensor data, mapping 24 input features to a single curvature value.}
    \label{fig:mlp_arch}
\end{figure}
For curvature prediction, we implemented a regression model based on a residual multilayer perceptron (MLP). Figure~\ref{fig:mlp_arch} shows the model architecture. The model takes a 24-dimensional feature vector as input, consisting of 16 normalized node readings from the tactile array and 8 engineered global features (sum, mean, standard deviation, minimum, maximum, range, $\ell_{2}$ norm, and interquartile range).  

The proposed network includes a stem block that expands the input to a 128-dimensional hidden space, followed by three residual blocks that refine this representation using fully connected layers with nonlinear activations and skip connections. The final regression head compresses the features to a single curvature output ($\kappa$) in m$^{-1}$. Dropout regularization and normalization are applied throughout to improve stability and generalization.

\vspace{-.2em}

\subsection{Model Training and Validation}

To build the curvature prediction dataset, three tactile sensors were mounted on ten cylindrical substrates with curvatures ranging from 0 to 80~m$^{-1}$. For each curvature, ten baseline measurements were collected for each sensor, each averaged over 100 samples to minimize noise. The sensors were rotated and flipped during data collection to introduce orientation and deformation variability, improving robustness against mounting differences.  

Each data sample contained 16 normalized node readings and 8 global statistical features, forming a 24-dimensional input vector. The dataset was divided into training (60\%), validation (20\%), and test (20\%) sets. The MLP was trained for 500 epochs using the AdamW optimizer with Huber loss and a cosine learning rate schedule. Two data augmentation methods were used to enhance generalization: Mixup, which blends pairs of samples to smooth input–output mappings, and label jitter, which adds small Gaussian noise to the targets to account for measurement uncertainty.  

\begin{figure}[b]
    \centering
    \includegraphics[width=0.7\linewidth]{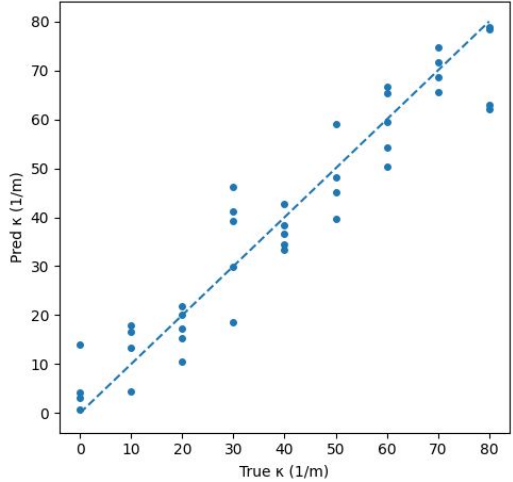}
    \caption{\textbf{MLP Training Result.} Prediction performance of the residual MLP on the test set, showing predicted versus true curvature values. Each curvature level includes five independent samples.}
    \label{fig:training_result}
\end{figure}

The residual MLP achieved an RMSE of $7.53\,\mathrm{m}^{-1}$, MAE of $5.90\,\mathrm{m}^{-1}$, and $R^2=0.91$ using the test set from $0$ to $80\,\text{m}^{-1}$. $100\,\text{m}^{-1}$ was excluded from curvature prediction training due to the unreliable response observed in Section \ref{sec:results}. These results indicate that the model can explain over 90\% of the variance in curvature across the tested range of $0$-$80\,\mathrm{m}^{-1}$. Figure~\ref{fig:training_result} shows the predicted versus true curvatures.

\section{Force Calibration} \label{sec:force_cali}

In this section, we present the force calibration procedure. We outline the experimental setup used to measure sensor response across surfaces with different curvatures and present the calibration model that related force to sensor output.

\subsection{Methods}

\begin{figure}[t]
    \centering
    \includegraphics[width=1.0\linewidth]{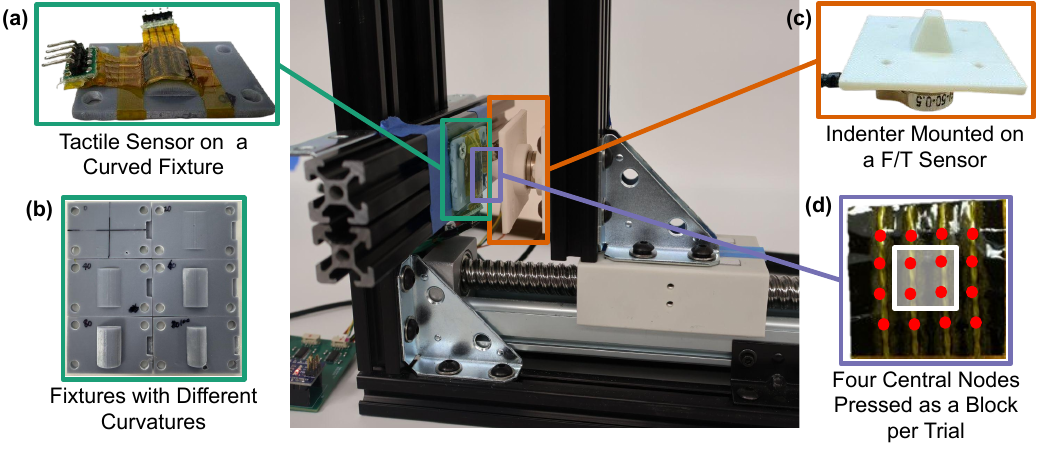}
    \caption{\textbf{Force Calibration Experimental Setup:} (a) Tactile sensor mounted on a curved fixture. (b) 3D printed fixtures with different curvatures. (c) Indenter mounted on a force/torch sensor to apply controlled loads. (d) In each trial, the indenter pressed the four central nodes simultaneously, forming a node block, to collect sensor responses under varying surface curvatures.}
    \label{fig:cal_setup}
\end{figure}

Figure~\ref{fig:cal_setup} shows the calibration system. A linear actuator driven by a stepper motor moves a 3D-printed indenter tip in a controlled manner. The tip is mounted on a force/torque (F/T) sensor (ATI Nano 17), which provides accurate ground-truth measurements of the normal force. Our $0.5\, \text{in}\times0.5\,\text{in}$, $4\times4$ sensor array is mounted to the 3D printed cylindrical substrates with curvatures ranging from $0$ to $100\,\text{m}^{-1}$ as used for the curvature prediction described in Section~\ref{sec:curv}.

During experiments, the indenter tip contacts the sensor generating two streams of data: (i) the \textit{normal force} from the F/T sensor, and (ii) the \textit{voltage} outputs from the 16 sensing nodes through our readout circuit. We perform calibration at the block level by grouping nodes together. We chose block-level calibration since individual nodes varied widely, while grouping reduced variability, produced more stable results, and has been used in prior work~\cite{huang_3d-vitac_2025}. 

\vspace{-.5em}

\subsection{Results} \label{sec:results}

\begin{figure}[]
    \centering
    \includegraphics[width=0.9\linewidth]{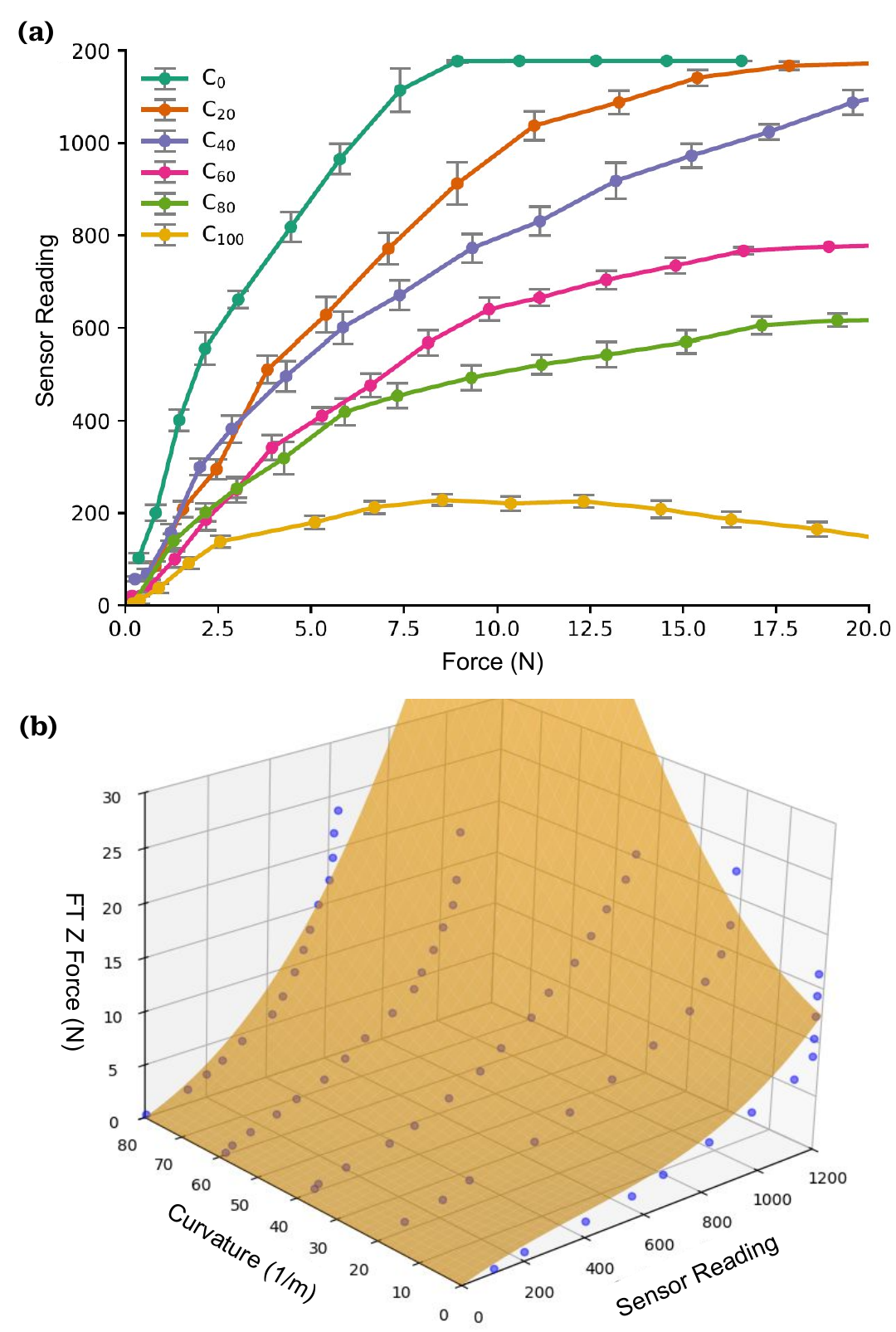}
    \caption{\textbf{Force Calibration Results.} (a) Sensor readings were measured under applied forces from 0-20 N on fixtures with varying curvatures ($C_0$-$C_{100}$). Each data point represents the mean$\pm$standard deviation of 100 consecutive sensor readings. Higher curvature was observed to reduce sensitivity and shift the saturation point. (b) A third-degree regression model was fitted to the experimental data to generate a calibration surface, illustrating the joint relationship between applied force, curvature, and sensor reading.}
    \label{fig:cal_result}
\end{figure}

Figure~\ref{fig:cal_result} shows the static calibration results of the sensor under different curvature conditions. 
The results show that as the surface curvature increases, the sensor becomes less sensitive to force: for the same applied force, the measured output is smaller, and the maximum attainable sensor value is also reduced. Moreover, the growth of the sensor reading with increasing force becomes slower on higher-curvature surfaces compared to flatter ones. This behavior arises because bending the sensor slightly stretches the piezoresistive layer, increasing the distance between conductive particles within the Velostat~\cite{yuan_velostat_2022}. As a result, fewer conductive pathways are available for current flow, which reduces the material’s responsiveness to pressure changes and lowers overall sensitivity~\cite{dzedzickis_polyethylene-carbon_2020, yuan_velostat_2022}.
At high curvatures approaching $C_{100}$, the sensor response fluctuates non-monotonically and exhibits minimal change with applied force, leading to decreased accuracy in force prediction. These findings confirm that while the sensor can cover a wide curvature range up to $C_{100}$, its reliability diminishes at the upper end, making curvature a critical factor to consider in calibration.

\subsection{Sensor Modeling}
To model the relationship between sensor reading and force, we performed polynomial regression using both the sensor output and curvature as inputs. We used a third-degree polynomial to fit the data, which captured the nonlinear dependencies of the sensor response. The resulting regression surface is shown in Fig.~\ref{fig:cal_result}. The model achieved an $R^2$ score of 0.9222 using the data from $0$ to $80\,\text{m}^{-1}$. The fitted mapping from sensor value ($S$) and curvature ($C$) to force ($F$) is given by: 
\[
F = 0.009625\,S - 0.000014\,S^{2} - 0.000372\,S\cdot C + 0.000005\,S\cdot C^{2}
\]

\section{Workflow Pipeline} 

\begin{figure}[t]
    \centering
   \includegraphics[width=\linewidth]{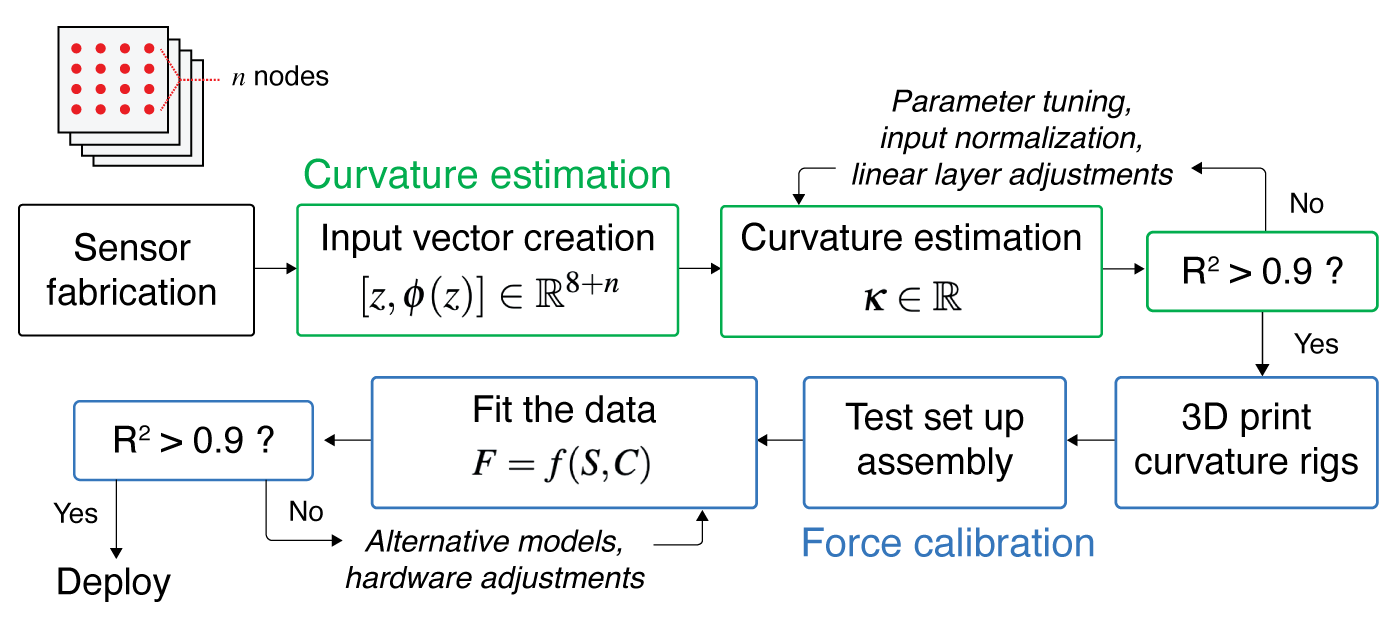}
    \caption{\textbf{Workflow Pipeline.} Overview of the calibration process, including curvature estimation (green) and force calibration (blue). The pipeline trains a curvature model, validates performance ($R^{2} > 0.9$), and then fits force as a function of sensor reading and predicted curvature.}
    \label{fig:placeholder}
\end{figure}

To promote reproducibility and adoption, we present a general pipeline for calibrating resistive tactile sensors (Fig.~\ref{fig:placeholder}). The process begins with fabricating a sensor and extracting baseline signals, which are combined with a compact feature representation to form the model input. An MLP then estimates surface curvature, which is iteratively validated until the model achieves sufficient accuracy (e.g., $R^{2} > 0.9$). Model performance can be further improved through hyperparameter tuning, normalization, or training across multiple sensors and curvatures. Once curvature estimation is reliable, controlled data collection is enabled using a reference force sensor, linear actuator, and 3D-printed fixtures spanning different curvatures. Calibration curves are then derived to map force to sensor readings as a function of curvature. If accuracy thresholds are not met, alternative models can be employed and the process repeated. The pipeline thus provides a modular framework that others can adapt and extend to their own tactile sensing platforms.


\section{Curvature-Calibrated Sensor Performance}


In this section, we evaluate the performance of the curvature-aware calibration model. We attach the sensor to five daily objects with varying surface curvatures to assess its ability to predict curvature and estimate force accurately. Each experiment includes both flat-surface and curvature-aware calibration for comparison. The results demonstrate how curvature affects sensing behavior and highlights the improvements achieved through curvature-aware modeling.

\begin{figure*}[!tbp]
    \centering
    \includegraphics[width=.9\linewidth]{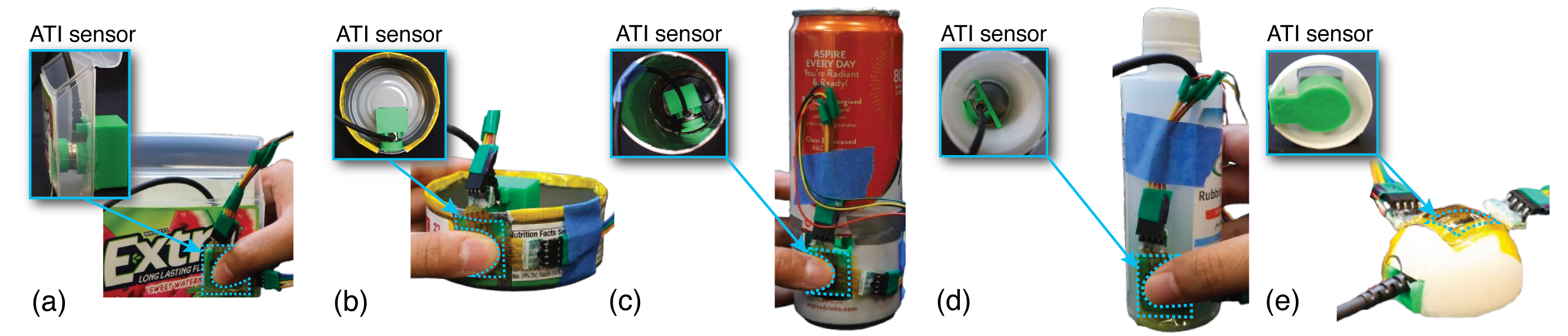}
    \caption{\textbf{Experimental Setup for Curvature Estimation of Everyday Objects.} Tactile sensor mounted on five daily objects with different surface curvatures for curvature and force evaluation. The ATI F/T sensor was embedded beneath the contact surface to provide ground-truth force measurements.}
    \label{fig:daily_objects}
\end{figure*}

\subsection{Method}

\begin{table*}[t]
\centering
\scriptsize
\renewcommand{\arraystretch}{1.1}
\setlength{\tabcolsep}{3pt}
\begin{tabular}{lccccccccccccc}
\toprule
\multirow{2}{*}{Object} & \multicolumn{2}{c}{Curvature (m$^{-1}$)}
& \multicolumn{2}{c}{2 N} & \multicolumn{2}{c}{4 N} & \multicolumn{2}{c}{6 N} & \multicolumn{2}{c}{8 N} & \multicolumn{3}{c}{Natural Holding} \\ 
\cmidrule(lr){2-3}\cmidrule(lr){4-5}\cmidrule(lr){6-7}\cmidrule(lr){8-9}\cmidrule(lr){10-11}\cmidrule(lr){12-14}
 & \emph{GT} & PR & Flat & Curve & Flat & Curve & Flat & Curve & Flat & Curve & \emph{GT} & Flat & Curve \\
\midrule
(a) Gum box          & 0.00  & 4.82  & 0.00$\pm$0.23 & 0.14$\pm$0.22 & 0.08$\pm$0.17 & 0.22$\pm$0.15 & 0.61$\pm$0.08 & 1.03$\pm$0.07 & 2.41$\pm$0.09 & 2.85$\pm$0.08 & 1.92$\pm$0.09 & 0.30$\pm$0.15 & 0.41$\pm$0.14 \\
(b) Tuna can         & 12.03 & 15.80 & 0.75$\pm$0.07 & 0.15$\pm$0.13 & 1.77$\pm$0.04 & 0.08$\pm$0.07 & 3.25$\pm$0.08 & 1.11$\pm$0.14 & 4.28$\pm$0.04 & 1.21$\pm$0.08 & 2.31$\pm$0.07 & 1.02$\pm$0.18 & 0.08$\pm$0.32 \\
(c) Energy drink     & 17.54 & 20.68 & 0.94$\pm$0.10 & 0.64$\pm$0.24 & 2.54$\pm$0.05 & 0.34$\pm$0.12 & 2.98$\pm$0.17 & 1.97$\pm$0.49 & 4.20$\pm$0.02 & 2.33$\pm$0.02 & 2.23$\pm$0.06 & 0.73$\pm$0.10 & 0.36$\pm$0.19 \\
(d) Rubbing alcohol     & 25.00 & 25.30 & 1.44$\pm$0.06 & 0.11$\pm$0.22 & 2.95$\pm$0.11 & 0.39$\pm$0.38 & 4.36$\pm$0.03 & 0.24$\pm$0.11 & 5.68$\pm$0.05 & 0.35$\pm$0.20 & 1.68$\pm$0.15 & 1.23$\pm$0.14 & 0.16$\pm$0.49 \\
(e) Ping pong ball   & 50.00 & 42.28 & 1.71$\pm$0.08 & 0.57$\pm$0.72 & 3.51$\pm$0.03 & 0.27$\pm$0.26 & 5.34$\pm$0.04 & 0.17$\pm$0.33 & 6.91$\pm$0.03 & 1.84$\pm$0.28 & 1.26$\pm$0.24 & 1.01$\pm$0.06 & 0.43$\pm$0.21 \\
\bottomrule
\end{tabular}
\caption{GT and PR denote ground-truth and predicted curvature ($m^{-1}$), respectively. Flat and Curve denote calibration methods. All reported values in Flat and Curve columns represent mean absolute force error with standard deviation $(\text{N} \pm \text{SD})$  obtained using flat and curvature-aware calibration.}
\label{tab:val_results}
\end{table*}


To evaluate the performance of the curvature-aware calibration model and the sensor’s curvature prediction capability under realistic conditions, five common daily objects with varying surface curvatures were selected. These objects spanned a curvature range from $0~\mathrm{m^{-1}}$ to $50~\mathrm{m^{-1}}$, representing flat, mildly curved, and moderately curved geometries.

Each object was instrumented with a F/T sensor (ATI Nano 17) embedded beneath the contact area to provide ground-truth force measurements. The tactile sensor was then attached to the object’s curved surface (Fig.~\ref{fig:daily_objects}). Before force testing, the MLP model was executed using the sensor’s baseline readings to predict the local surface curvature. This predicted curvature was later used as an input to the curvature-aware calibration model.

A graphical user interface (GUI) was developed to standardize data collection and enable static force measurements through natural human interaction. The GUI displayed four sequential reference forces (2, 4, 6, and 8~N), corresponding to the typical fingertip grasp range during daily activities~\cite{rajakumar2022evidence}. Real-time plots of the F/T sensor readings and the reference force were presented on screen. During testing, the experimenter manually pressed the tactile sensor mounted on the object until the measured ground-truth force reached the target level. Once the force remained within $\pm0.2$~N of the reference value for 5s continuously, the system automatically advanced to the next reference force. The mean tactile sensor output over that 5s window was recorded as the steady-state response for that reference force level. 
In addition to the four reference forces, the experimenter’s natural holding force for each object was also measured. Both the F/T ground-truth data and the sensor readings were recorded for this condition.

Two calibration strategies were compared: (1) flat-surface calibration, a baseline model derived from flat-surface characterization, and (2) curvature-aware calibration, the proposed model that incorporates curvature as an input for improved force estimation.

\vspace{-.3em}

\subsection{Results}
Table \ref{tab:val_results} summarizes the results of curvature and force estimation across five daily objects. Only the gum box (Fig. \ref{fig:daily_objects} (a)) featured a flat surface, while the other objects exhibited various curvatures. Our model accurately predicted surface curvature, with a mean absolute error of 3.95 m$^{-1}$ across all five objects. The accuracy of force estimation depended on object geometry. For flat objects such as the gum box, both flat and curvature-aware calibration methods produced comparable force predictions. In contrast, for objects with curved surfaces, the flat-surface calibration consistently underestimated the applied force. This effect became more pronounced as the surface curvature increased. Across all objects and force levels, the flat calibration yielded higher force errors compared to the curvature-aware model. 

\vspace{-.3em}

\section{Discussion and Conclusion}

In this paper, we investigate the effect of surface curvature on the readings of flexible tactile sensors and demonstrate that curvature significantly alters sensing behavior. To address this challenge, we develop a curvature-aware calibration model that improves accuracy on non-flat surfaces and show that curvature itself can be inferred directly from sensor readings using an MLP. We introduce a workflow pipeline that combines accessible calibration methods with MLP training. Finally, we validate the practicality of our sensor and the robustness of the calibration model using everyday objects with diverse curvatures.

The findings from the performance test show that both the curvature and the applied force impact the error rates of our sensor. High curvatures tend to increase measurement errors, which is consistent with force calibration studies from Section~\ref{sec:force_cali}. Similarly, as greater force is applied, error rates in calibrated readings also tend to increase. This effect can be also attributed to Velostat's nonlinear resistance response to pressure and its inherent material properties~\cite{dzedzickis_polyethylene-carbon_2020}. Thus, future work should explore sensor designs with differing materials.

While this work focused primarily on 1D curvature surfaces and static calibration, future work will 
extend the framework to 2D curvature and full 3D mesh models for realistic deployment on irregular geometries. 
Addtionally, we aim to advance from static to dynamic calibration so the sensor can adapt in real time and be used for versatile tactile sensing. 

\bibliographystyle{IEEEtran}
\bibliography{references}

\end{document}